\documentclass[a4paper, 10pt, conference]{ieeeconf}      
\usepackage{FG2021}

\FGfinalcopy 

\IEEEoverridecommandlockouts                              
\usepackage{graphicx} 
\usepackage{hyperref}

\usepackage{color}

\def\FGPaperID{****} 

\title{\LARGE \bf
 Safe Fakes: Evaluating Face Anonymizers for Face Detectors
}

\author{\parbox{16cm}{\centering
    {\large Sander R. Klomp$^{12}$, Matthew van Rijn$^3$, Rob G.J. Wijnhoven$^2$,\\ Cees G.M. Snoek$^3$, Peter H.N. de With$^1$}\\
    {\normalsize
    $^1$ Department of Electrical Engineering, Eindhoven University of Technology, Eindhoven, The Netherlands\\
    $^2$ ViNotion B.V., Eindhoven, The Netherlands\\
    $^3$ Informatics Institute, University of Amsterdam, Amsterdam, The Netherlands}}
}

\begin{document}

\ifFGfinal
\thispagestyle{empty}
\pagestyle{empty}
\else
\author{Anonymous FG2021 submission\\ Paper ID \FGPaperID \\}
\pagestyle{plain}
\fi
\maketitle

\begin{abstract}
Since the introduction of the GDPR and CCPA legislation, both public and private facial image datasets are increasingly scrutinized. Several datasets have been taken offline completely and some have been anonymized. However, it is unclear how anonymization impacts face detection performance. To our knowledge, this paper presents the first empirical study on the effect of image anonymization on supervised training of face detectors. 
We compare conventional face anonymizers with three state-of-the-art Generative Adversarial Network-based (GAN) methods, by training an off-the-shelf face detector on anonymized data. Our experiments investigate the suitability of anonymization methods for maintaining face detector performance, the effect of detectors overtraining on anonymization artefacts, dataset size for training an anonymizer, and the effect of training time of anonymization GANs. A final experiment investigates the correlation between common GAN evaluation metrics and the performance of a trained face detector.
Although all tested anonymization methods lower the performance of trained face detectors, faces anonymized using GANs cause far smaller performance degradation than conventional methods.
As the most important finding, the best-performing GAN, DeepPrivacy, removes identifiable faces for a face detector trained on anonymized data, resulting in a modest decrease from 91.0 to 88.3 mAP. 
In the last few years, there have been rapid improvements in realism of GAN-generated faces. We expect that further progression in GAN research will allow the use of Deep Fake technology for privacy-preserving Safe Fakes, without any performance degradation for training face detectors.
\end{abstract}

\begin{figure}[t]
    \centering
    \includegraphics[width=\linewidth]{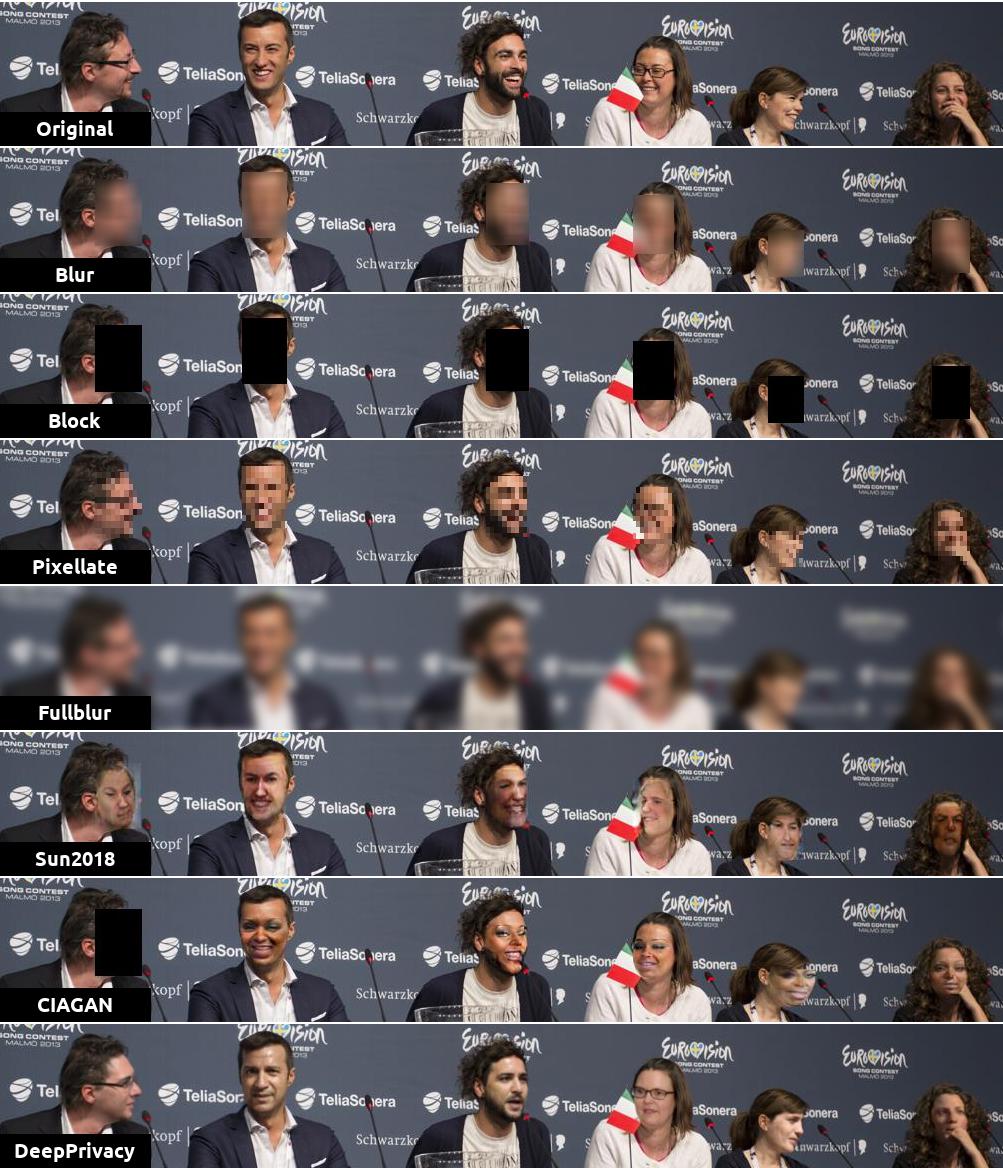}
    \caption{Example images as anonymized by all methods tested in this paper. We evaluate the performance of a state-of-the-art face detector trained on the anonymized faces for each method, assess their generalizability and analyze the best anonymized face detector setting in more detail. DeepPrivacy generates most realistic faces, achieving anonymization with the least artefacts. Best viewed digital and zoomed in.}
    \label{fig:anonymization-examples}
\end{figure}

\section{Introduction}
Since the introduction of the GDPR and CCPA, the use of privacy information in public academic datasets has become a serious topic of debate in science and society. For this reason, several datasets, such as the well-known DukeMT dataset~\cite{Ristani2016}, 80 million tiny images~\cite{Torralba2008} and MS Celeb faces dataset~\cite{Guo2016} have all been taken offline~\cite{Satisky2019, Murgia2019}. Recently, the creators of the most famous ImageNet~\cite{JiaDeng2009} decided to blur all faces in their dataset for privacy protection~\cite{Yang2021}. 
To prevent the loss of academic, private and commercial datasets due to privacy legislation, datasets should be anonymized wherever possible. For this reason, this paper focuses on one of the worst-case scenarios for deep learning on anonymized data: anonymization of face detection datasets and its impact on training face detectors. Examples of the anonymization methods that are tested are shown in Fig.~\ref{fig:anonymization-examples}. Knowing which anonymization method removes identifiable information from face detection datasets, while preserving important facial features, would be of huge benefit to those who want to be at the safe side of privacy laws and respect and honor the rights of those depicted. This would also allow new datasets to be more easily created and published, while  existing datasets can be safely anonymized. Thus, our objective is to find a face anonymization that enables the use of faces in training data, but prohibits the identification of individual subjects.

One of the regulations that restricts the use of privacy-sensitive datasets is the GDPR. The GDPR stipulates that personal data that is not anonymized, including faces, should not be processed or stored without explicit consent, unless it is ``strictly necessary" for the application~\cite{voigt2017eu}.
Yang~\textit{et al.}~\cite{Yang2021} conclude that for ImageNet, training deep neural networks with blurred faces results in less than 1\% performance degradation on average for the ImageNet classes, with a significant drop of up to 9\% for classes that are related to faces, such as ``harmonica". However, ImageNet does not have ``face" labels, so it is not directly suitable to evaluate the impact of anonymizing faces on face detection performance.
Therefore, in this paper, we investigate the use of anonymization techniques for anonymizing face detection datasets, which contain vastly more privacy-sensitive images with recognizable faces than ImageNet and where intuitively the faces are ``strictly necessary" for training a face detector. 

There is already a wealth of face anonymization methods~\cite{padilla2015visual}. Simple anonymizers such as blurring are commonly used on images taken in public. There are also more advanced anonymizers that replace faces with completely new ones generated with Generative Adversarial Networks (GANs), e.g.,~\cite{goodfellow2014generative, sun2018natural, hukkelaas2019deepprivacy, maximov2020ciagan}. Fig.~\ref{fig:anonymization-examples} shows faces generated by some state-of-the-art GANs.

In this paper, we consider face detector training, after applying face anonymization as a preprocessing step.
Whereas the goal of GAN-based anonymization methods is to generate a visually pleasing result, they do not necessarily preserve the variety of features important for training a face detector and possibly introduce artefacts. CIAGAN~\cite{maximov2020ciagan} and DeepPrivacy~\cite{hukkelaas2019deepprivacy} test how well a pretrained face detector can detect their generated faces, but this does not guarantee the suitability of the generated faces for detector training. A face with some minor artefacts is likely still detectable by a pretrained face detector, but conversely, a training set in which every face has the same artefacts will produce a detector that struggles to detect real faces, as will be shown in this paper.

This paper makes three contributions.
\begin{itemize}
    \item To our knowledge, we are the first to study the effect of image anonymization on the training of face detectors.
    \item Besides conventional face anonymization methods, we also evaluate the face hallucinating abilities of GANs on face detection training performance.
    \item We analyze the impact of detector training time, anonymizer training time and dataset size on detector performance. We also determine which common GAN evaluation metric correlates best with the detection performance.
\end{itemize}


\section{Related work}

\subsection{Face detection} \label{sec_related_work_subsec_face_detection}
Face detection has gone through several periods, but today deep convolutional neural networks (DCNNs) are dominant. As face detection is a subproblem of generic object detection, face detectors generally use object detection networks as a starting point, such as the widely used Faster-RCNN~\cite{ren2015faster} and SSD~\cite{liu2016ssd}.
Detecting small faces is generally considered challenging, hence the Tiny Face detector~\cite{hu2017finding} and Dual-Shot Face Detector (DSFD)~\cite{li2019dsfd} focus on improving the detection of very small faces by exploiting multiple scales and a feature enhance module, respectively. 
For the popular face detection dataset WIDER FACE~\cite{yang2016wider}, a leaderboard\footnote{\url{http://shuoyang1213.me/WIDERFACE/WiderFace_Results.html}} 
is maintained showing the top scoring face detectors. At the time of writing it is led by the Automatic and Scalable Face Detector (ASFD)~\cite{zhang2020asfd}, though the performance difference with other top detectors~\cite{deng2019retinaface, earp2019face, zhang2019accurate, li2019dsfd} is small.

For our experiments there are two requirements for the face detector. First of all, the detector should converge in a reasonable number of iterations to allow for experiments in different settings. Second, the network should achieve performance that is close to state-of-the-art, because the most difficult faces to detect are likely also the most difficult to anonymize well. The Dual-Shot Face Detector (DSFD)~\cite{li2019dsfd} satisfies these requirements with near-state-of-the-art performance on the WIDER FACE dataset within a reasonable number of 60,000 training iterations.

\subsection{Face anonymization}
Conventional face anonymization or ``de-identification" research has long focused on techniques such as blurring, blocking and pixelation~\cite{ribaric2016identification}. These methods are simple to apply and deploy at a large scale. Since face detectors learn to detect faces based on the same features that are obfuscated by these early anonymization methods, it is unlikely that blurring, blocking or pixelation allow face detectors to learn useful features. Furthermore, blurring and pixelation do not always prevent machine identification~\cite{sun2018natural, ribaric2016identification} and methods to deblur images have also been explored~\cite{kundur1996blind}.

Newer face anonymization methods have appeared, able to replace the faces in an image with entirely different faces. These face-replacement methods improve upon the early anonymization methods in two ways. Firstly, by replacing the original face instead of merely deforming it, a greater level of privacy protection is achieved. Secondly, the faces look more natural, which is a benefit if they are meant for display to humans, or if the features are needed to train a face detector. An early replacement-based anonymization method is the $k$-Same algorithm~\cite{newton2005preserving}. This method replaces the face with the average of the $k$-closest faces when projecting the faces to an eigenspace. While effective, a drawback of $k$-Same is that all faces must be aligned to be able to calculate distances between them and obtain average faces.

Recently, deep convolutional neural networks have also become successful in face anonymization. The biggest change is the advent of the Generative Adversarial Network (GAN)~\cite{goodfellow2014generative} and its variants. These neural networks are able to generate new faces from noise or context, alleviating major privacy concerns. State-of-the-art GANs for face generation such as~\cite{karras2019style} produce very detailed high-resolution faces. These faces are not suitable to be used to anonymize faces within a larger image, because they do not blend the faces with the background in the original image. Dedicated face anonymization GANs such as CIAGAN~\cite{maximov2020ciagan}, DeepPrivacy~\cite{hukkelaas2019deepprivacy}, CLEANIR~\cite{Cho2020} and~\cite{sun2018natural, sun2018hybrid}, solve this issue by generating new faces from the area around the face instead of random noise, similar to inpainting. As is common for inpainting, the generator often consists of an autoencoder such as U-Net~\cite{Ronneberger2015}. The GAN's discriminator then judges the generated face and the area around it. This results in properly blended faces. 

In addition to using an autoencoder as the generator, another method shared between many GAN-based face anonymizers~\cite{maximov2020ciagan, hukkelaas2019deepprivacy, sun2018natural} is the use of landmarks to represent the positions of certain parts of the face. Structure guidance through landmarks can improve the quality of complex structures in generated images~\cite{ma2017pose}. Since landmarks determine the facial structure, they can be used to preserve the pose of the original face.

In this work we investigate the impact of both the conventional methods and GAN-based face anonymizers. CIAGAN and DeepPrivacy investigate whether their anonymized faces can be detected using a detector trained on real faces. However, being able to detect an anonymized face does not guarantee that the face is suited for training a detector. Thus, we instead investigate the impact of anonymizing the training data itself and what the impact is on a detector trained on this data.  Our experiments are inspired by the recent anonymization of ImageNet~\cite{Yang2021}, but whereas they only blur faces to measure impact on their general object classification task, we consider face anonymization for the particular problem of face detection. Hence, in our setting faces are the target class, instead of (accidental) background context for other classes.

\begin{figure}[t]
    \centering
    \includegraphics[width=\linewidth]{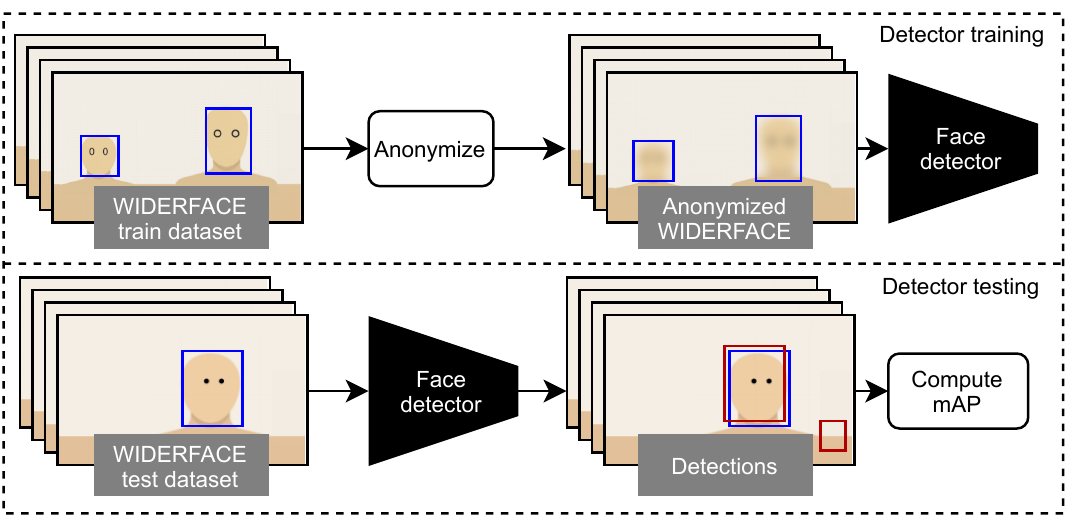}
    \caption{Schematic of the face detector testing framework. We investigate the challenging case of anonymizing training data, not testing data.}
    \label{fig:DetectorTrainingAndTesting}
\end{figure}

\begin{figure}[t]
    \centering
    \includegraphics[width=\linewidth]{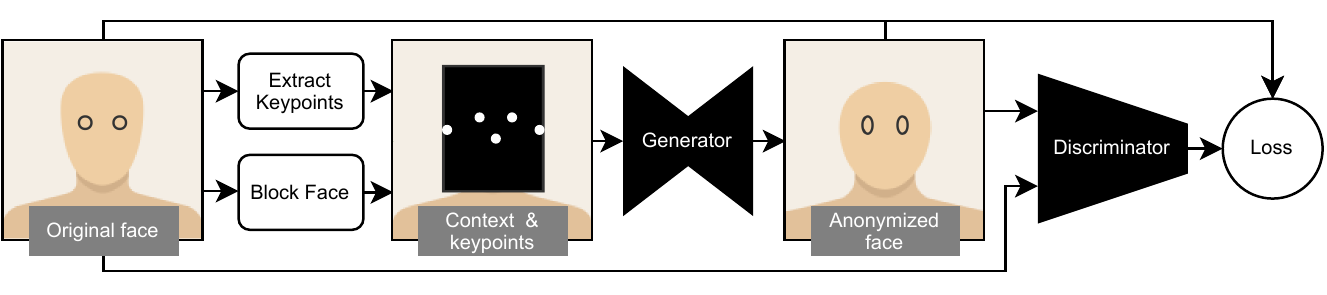}
    \caption{General training schedule of face anonymization GANs.}
    \label{fig:AnonymizerTraining}
    \vspace{-4mm}
\end{figure}

\section{Safe Fakes Experimentation Protocol}
For our experimentation protocol, we rely on basic anonymization techniques
as a point of reference: blurring, blocking and pixelation. Although Yang~\textit{et al.}~\cite{Yang2021} concluded that blurring of faces in ImageNet has a minimal impact on training general object classifiers, we deem it unlikely that these simple anonymization methods will not harm the training of face detectors. We consider GAN-based anonymizations that generate realistic faces more promising.

This brings us to the key research question of our work: ``Can we anonymize faces while maintaining effective training of face detectors?". To answer this question, we design several experiments. A schematic representation of our experimental setup is shown in Fig.~\ref{fig:DetectorTrainingAndTesting}. First, the faces in the training set of the WIDER FACE~\cite{yang2016wider} are anonymized using all the methods shown in Fig.~\ref{fig:anonymization-examples}. 
This fully anonymized dataset is then used to train the DSFD~\cite{li2019dsfd} face detector in the exact same way it would normally be trained on original data. To test the detector, the mAP score is computed on the normal, not anonymized, validation set. This setup is realistic for real-world applications, such as surveillance or autonomous driving. Here the training dataset must be stored (and thus should be anonymized if possible) for later potential improvements to the detector. In contrast, the testing dataset simulates live deployment on newly captured images, where the input faces are not anonymized.

\subsection{Face anonymizers}
Anonymization is applied to every face bounding box in the dataset. A fixed blur kernel size is employed regardless of bounding box size, pixelation is set to a fixed output size, and GAN methods scale cropped faces to a fixed input resolution.
Besides the training of the face detector, the GAN-based anonymization methods also need to be trained. We focus on three recent GAN-based anonymization networks: the work of Sun~\textit{et al.}~\cite{sun2018natural} (Sun2018), CIAGAN~\cite{maximov2020ciagan} and DeepPrivacy~\cite{hukkelaas2019deepprivacy}. The employed general training setup of these methods is shown in Fig.~\ref{fig:AnonymizerTraining}. For CIAGAN and DeepPrivacy, publicly available code is used, but Sun2018 does not have a public implementation available, so we implement it from scratch. All three methods require the extraction of keypoints from an original face and then black out the face to prevent leakage of privacy-sensitive data. The generator inpaints a face based on the keypoints and context area, which is judged by the discriminator. Both the datasets on which the anonymizers are trained and the implementation details differ substantially between the tested methods. However, in order to perform a decent scientific evaluation, we attempt to harmonize the implementation details to enable a fair comparison, as explained next.

\setlength{\tabcolsep}{5pt} 
\renewcommand{\arraystretch}{1.2} 
\begin{table}[t]
\caption{Basic comparison of GAN-based anonymization methods. Conventional methods are discussed in text.}
\label{tab:GAN-comparison}
\centering
\begin{tabular}{lccc}
\hline
\textbf{Anonymizer}  & \textbf{Keypoints} & \textbf{Input blocking}          & \textbf{Face resolution} \\ \hline
Sun2018~\cite{sun2018natural}    & 68        & Black / Blurred box & $256\times256$       \\
DeepPrivacy~\cite{hukkelaas2019deepprivacy} & 7         & Black / Noise box   & $128\times128$        \\
CIAGAN~\cite{maximov2020ciagan}   & 27   & Black mask              & $128\times128$        \\ \hline
\end{tabular}
\vspace{-4mm}
\end{table}

\subsubsection{Implementation details for fair comparison of GANs}
Some differences between the GAN-based anonymization networks make it hard to compare them fairly. A summary of these key differences is shown in Table~\ref{tab:GAN-comparison}.

\textbf{Keypoints} Both Sun2018 and CIAGAN use the Dlib facial keypoint detector to detect 68 keypoints, of which CIAGAN only uses 27. DeepPrivacy employs only seven keypoints detected using Mask-RCNN~\cite{He}, two of which are on the shoulders, not on the face. The seven keypoints of DeepPrivacy can be reliably detected on very small faces, as low as 10 pixels wide, at which point the face is not recognizable anymore anyway. The 68-keypoint Dlib detector limits Sun2018 and CIAGAN to larger faces of at least around 50 pixels wide. Without detected keypoints, CIAGAN cannot anonymize faces at all, while Sun2018 still generates some face-like blobs.

\textbf{Input blocking.} In all our experiments we use black-out input blocking, as it is supported by all three methods and thus most fair. Blurred box inputs for Sun2018 improve detector training performance slightly ($\approx$3\%), but generated faces more often resemble the original, reducing obfuscation effectiveness. CIAGAN does not black out the entire bounding box, but only a mask of the face based on the detected keypoints. This gives it an advantage in that the network has to learn only very limited inpainting, but prevents faces from being generated if keypoints are not available.

\textbf{Face resolution.} DeepPrivacy and CIAGAN generate faces at a resolution of $128\times128$ pixels while Sun2018 uses $256\times256$. However, we see less artefacts in our re-implementation of Sun2018 when we use $128\times128$ pixels, hence this same resolution is used for all networks in all experiments.

\textbf{Skip connections.}  All three methods rely on U-Net~\cite{Ronneberger2015} generator with a symmetrically sized encoder and decoder. Both DeepPrivacy and Sun2018 employ skip connections to improve facial detail, while CIAGAN does not. Adding skip connections to CIAGAN would reduce the effectiveness of its identity selector input due to information leakage, so we leave it as-is.


\textbf{Training parameters.} All experiments are performed with the parameters of the original papers, wherever possible. Due to memory constraints, batch size and learning rate are reduced. For Sun2018, we use a batch size of 64, a learning rate of 5e-4 for the generator and 2e-5 for the discriminator. For DeepPrivacy we use a batch size of 128 and learning rate of 1.5e-3 and finally for CIAGAN we use a batch size of 16 and learning rate of 1e-4. As a face detector, the VGG-16 backbone variant of DSFD is employed~\cite{li2019dsfd} with a batch size of 4 and learning rate of 2.5e-4.

\subsection{Datasets}
We consider two types of datasets for our experiments: those for training the face anonymizer and those for training and testing the detector. The datasets for training the anonymizer and the detector should be different for two reasons. First, face detectors and face anonymizers have different requirements. Anonymizers require high-resolution faces to be able to generate faces at any scale, while face detection is often performed on lower resolution faces, especially in surveillance. Second, using the same dataset creates the risk of identifying information leaking into the anonymized dataset through the generator of the anonymizer, because it can simply memorize the identifiable faces.

\textbf{Face detector training and testing data}
For the face detector, the WIDER FACE dataset~\cite{yang2016wider} is used. This dataset is the logical choice for training the face detector because it is widely used, large and contains a great variety of different settings and face sizes. Additionally, WIDER FACE labels their faces with a difficulty label ``easy", ``medium" or ``hard", based on how hard the face is to detect for a face detector. This difficulty score correlates closely with properties such as occlusion, strange poses and face size, whereas smaller faces are considered harder to detect. Anonymization GANs are likely poor at anonymizing faces with strange poses and occlusion, which makes the ``hard" validation sets especially interesting to test on. The detector is trained on the (anonymized) training set and tested on the unaltered validation set.

\textbf{Face anonymizer data}
The applied datasets for training the anonymizers vary per experiment. In the first general comparison, we train the anonymizers on the datasets used in the original papers. A second comparison trains on just one dataset for all anonymizers, PIPA~\cite{piper}, for a more fair comparison. The PIPA dataset is chosen because it contains a good variety in faces of sufficient resolution, it has faces in realistic contexts and it is small enough that anonymizers converge quickly. In contrast to face detectors, anonymizers are always trained on crops of single faces.

The original datasets that the GANs are designed for are as follows.
CIAGAN is trained using a CelebA~\cite{liu2015faceattributes} subset of 1,200 identities with at least 30 faces each, resulting in a total of 40,000 faces. To anonymize an image, CIAGAN requires an identity choice as an input, which we select at random.
Sun2018 is originally trained on a custom pruned train/test split of the PIPA dataset of 34,000 faces. This dataset split is not available. In our re-implementation, we only use faces from the training and validation set for which the public Dlib
%
facial keypoint detector can detect landmarks, resulting in 16,000 faces.
DeepPrivacy is trained using the FDF~\cite{hukkelaas2019deepprivacy} dataset of 1.47M faces, which is a subset of the YFCC-100M~\cite{Thomee_2016} dataset.

\subsection{Experimental setup}
All experiments are performed on a single RTX 2080 Ti.
For the anonymization of the training sets we attempt to apply each anonymization method to every annotated face in the WIDER FACE training set. The conventional methods, blurring, black out and pixelation, can trivially be applied to the annotated bounding box. Blurring can also be applied to an entire image. We use a fixed kernel size of $25\times25$ for individual face blurring, a kernel size of $18\times18$ for full image blurring and a pixelation output size of $10\times10$ pixels. For the GANs, it is not always possible to anonymize the face due to a lack of detectable keypoints. Faces smaller than 10 pixels wide are not anonymized, but they are typically not recognizable. For larger faces the approach varies per network. For DeepPrivacy and Sun2018, if no keypoints can be detected, we run the network without keypoints, resulting in poor quality face-like blobs. Note that for DeepPrivacy, this case is rare, because keypoints are usually detected on faces larger than 10 pixels wide. For CIAGAN, if no keypoints can be detected, the face bounding box is blacked out. Examples of all anonymization methods applied to an image from WIDER FACE, are shown in Fig.~\ref{fig:anonymization-examples}.

\begin{figure*}[t]
    \centering
    \includegraphics[width=\linewidth,height=1.2in]{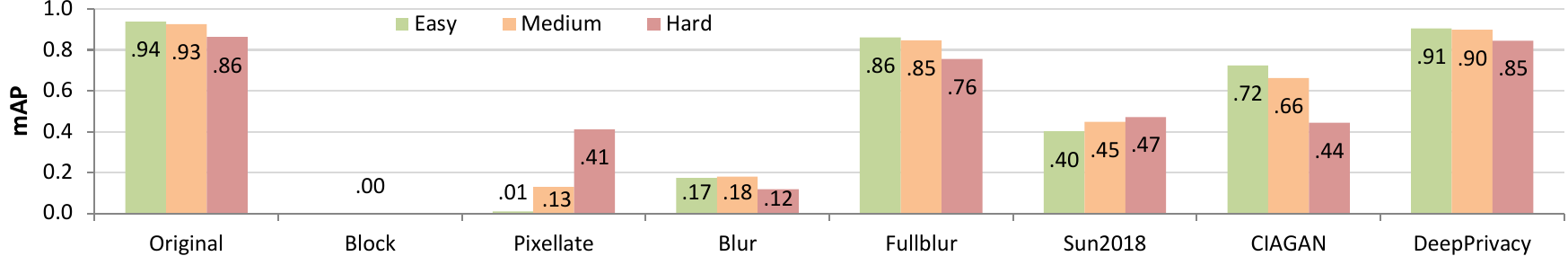}
    \caption{\textbf{Which face anonymizer performs best?} DSFD mAP scores on WIDER FACE validation set, per difficulty after being trained on the anonymized training set using different methods. DeepPrivacy performs best, with only a 1-3\% mAP score drop depending on face difficulty.}
    \label{fig:general-comparison}
    \vspace{-2mm}
\end{figure*}

\section{Experiments} \label{sec:experiments}
All experiments are aimed at answering the question ``Can we anonymize faces while maintaining effective training of face detectors?". Each subsection describes a single experiment that attempts to answer the question posed in its title. 

\subsection{Which face anonymizer performs best?}
\textbf{Setup.} In this experiment, the DSFD face detector is trained for the full 60,000 iterations on the WIDER FACE training dataset, as anonymized by either conventional or GAN-based methods.

\textbf{Quantitative results.} The results are shown in Fig.~\ref{fig:general-comparison} for each difficulty in the WIDER FACE dataset. Face blurring, blocking or pixelation turns the face detector into a blur-box, black-box or pixelated-box detector, respectively, instead of a face detector. This lowers the mAP considerably. Thus, the simple blurring approach performed on ImageNet~\cite{Yang2021} that proved effective for their image classification task is unsuited for face detection. 
Pixelation scores reasonably well on hard faces, because pixelation is fixed to $10\times10$ output pixels and hard faces are generally very small, thereby limiting the influence of pixelation. Blurring the entire image (Fullblur) yields a surprisingly small performance degradation, even for the small hard faces. 
GAN-based methods achieve better results, with DSFD trained on DeepPrivacy-anonymized faces reducing the performance by only 2.7\% mAP, compared to the unaltered dataset. 

\begin{figure}[t]
    \centering
    \includegraphics[width=\linewidth]{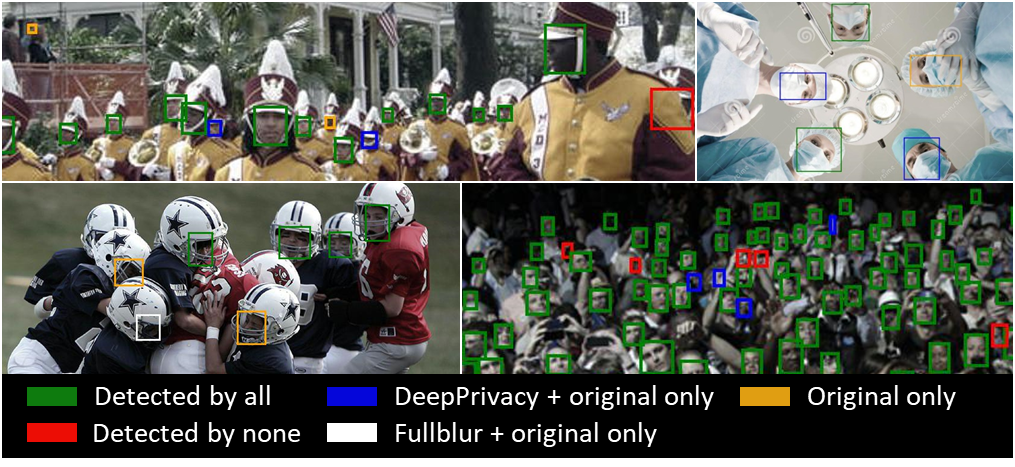}
    \caption{\textbf{Which face anonymizer performs best?} Qualitative comparison of detector trained on original, DeepPrivacy and Fullblur. Box color indicates the detector that could detect the face. Detection of heavily occluded, strange-pose and tiny faces suffer from anonymization. With the best anonymization method, DeepPrivacy, faces in white/orange are missed.}
    \label{fig:detection-examples}
    \vspace{-3mm}
\end{figure}

\textbf{Qualitative results.} A qualitative comparison of detections achieved using detectors trained on data anonymized through either Fullblur or DeepPrivacy is shown in Fig.~\ref{fig:detection-examples}. Compared to training on original data, training on anonymized images impacts faces of strange poses or strong occlusion the most. Qualitatively, DeepPrivacy outperforms Fullblur most noticeably on very small faces. Overall, this experiment shows that conventional methods that work on individual faces are not feasible, but DeepPrivacy and Fullblur show promising results. Note that for Fullblur, the degree of anonymization is strongly dependent on the face size, while DeepPrivacy hallucinates completely new faces regardless their size.

\subsection{What is the influence of face detector quality?}
\textbf{Setup.} The quality of a detector can be varied by changing the training time. In this experiment, the performance trend of increasingly powerful detectors is estimated by training the DSFD over three different training cycles of 7,500, 15,000 and 60,000 iterations. We hypothesize that this performance trend is likely to generalize to better detectors that will be developed in the future.

\begin{figure*}[t]
    \centering
    \includegraphics[width=0.95\linewidth]{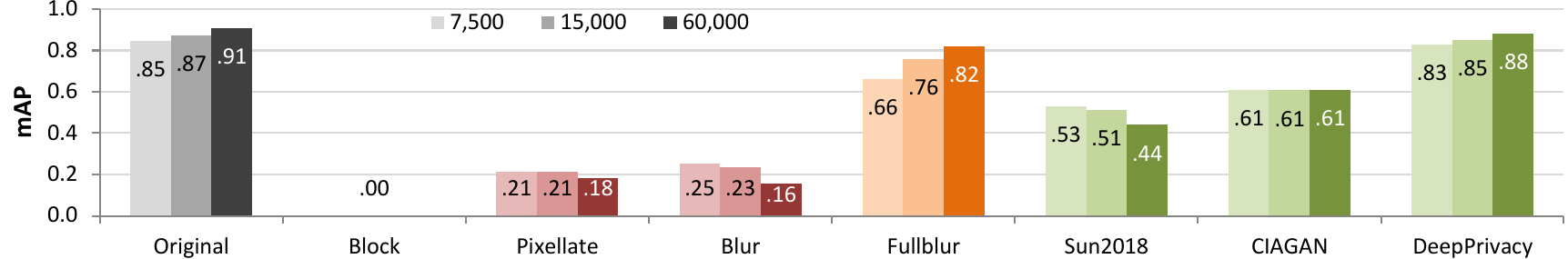}
    \caption{\textbf{What is the influence of face detector quality?} Detector training iterations versus mAP score. GANs shown in green, conventional in red, full-image blurring in orange. Anonymization methods introduce artefacts, on which the detector will overtrain, with notable exception of DeepPrivacy.}
    \label{fig:detector-training-time}
    \vspace{-2mm}
\end{figure*}

\textbf{Quantitative results.} The resulting performance trends are shown in Fig.~\ref{fig:detector-training-time}. Conventional methods that anonymize individual faces show a decreasing performance trend, indicating that the detector is likely overtraining on the anonymization artefacts. For the GAN-based methods, the trend is quite diverse. Our implementation of Sun2018 generates faces with visually the most severe blending artefacts compared to the other two GAN-based methods. Apparently, these artefacts are severe enough for the face detector to overtrain on them, since the performance of the detector drops with more iterations. The performance of CIAGAN is slightly better, because its masking strategy leads to far fewer blending artefacts. Interestingly, it appears that the generated faces are just realistic enough, so that the detector cannot overtrain on the artefacts. However, training performance is poor because the Dlib keypoint detector could not detect keypoints for all small faces. These keypoints are essential to CIAGAN's masking strategy. This causes a detector trained on CIAGAN-anonymized faces to only perform reasonably well on high-resolution faces and poorly on low-resolution ones (also visible in Fig.~\ref{fig:general-comparison}). For this reason, training longer slightly improves the performance on easy faces and lowers performance on hard faces, which balances to a constant score trend seen in Fig.~\ref{fig:detector-training-time}. Finally, DeepPrivacy performs best overall. It leaves most faces in the dataset suited for training, even small faces. Furthermore, detector performance improves when training for more iterations, suggesting that the faces that are generated are realistic enough so that the face detector learns to detect faces, rather than focus on anonymization artefacts. However, it should be noted that the relative performance loss of using DeepPrivacy, compared to training with original data, increases with more iterations of detector training, from a mere 1.6\% mAP at 7,500 iterations to 2.7\% mAP at 60,000 iterations. A possible explanation is that, qualitatively, the variety of faces generated by DeepPrivacy appears to be lower than in the original data. This causes faces in the long tail of the faces distribution to be detected less effectively. This suggests that when better face detectors are created by the research community, the performance gap is likely to grow, unless better anonymization networks are created as well.

\subsection{Why DeepPrivacy performs best?}
Previous experiments show DeepPrivacy is the most suitable method for anonymizing training data, but do not explicate why. Our hypothesis is that the primary cause is the large size of the dataset that DeepPrivacy is trained on (1.47M faces), compared to the 40,000 and 16,000 faces that CIAGAN and Sun2018 use. 

\begin{figure}[t]
    \centering
    \includegraphics[width=0.95\linewidth]{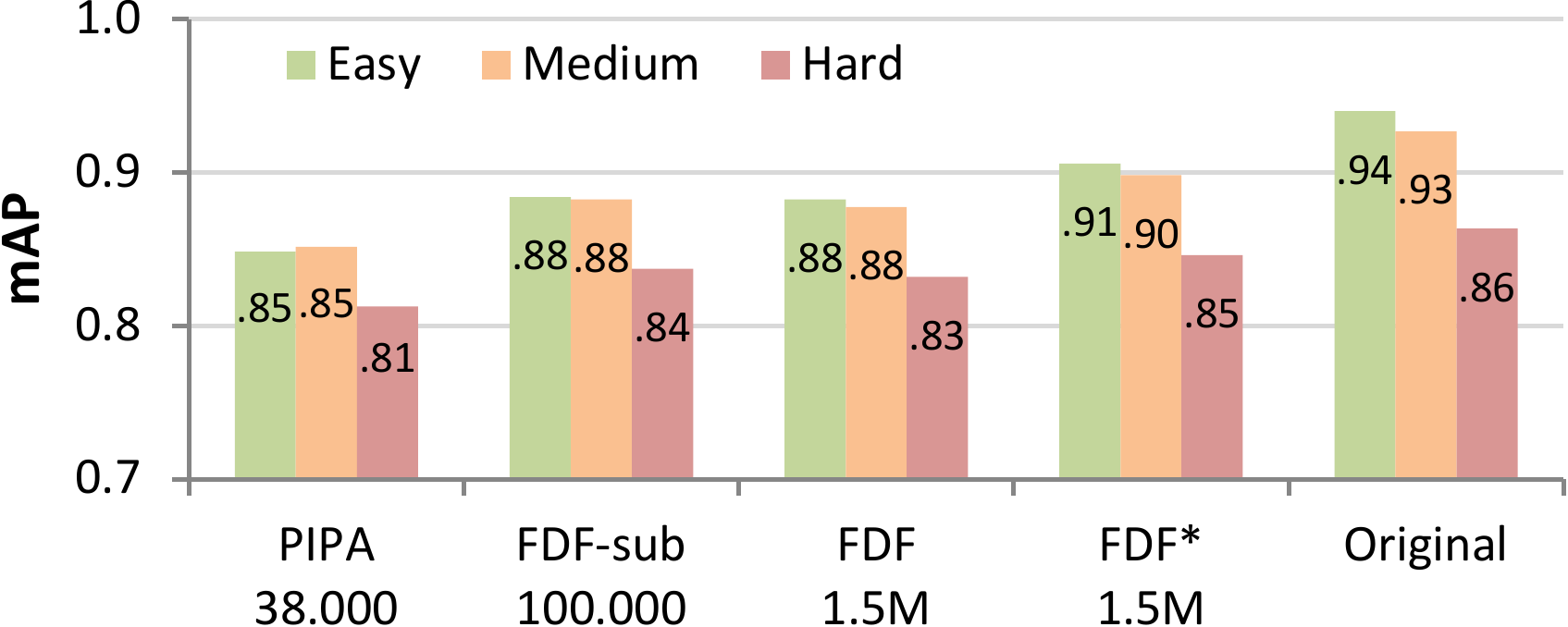}
    \caption{\textbf{Why DeepPrivacy performs best?} DSFD performance for increasing DeepPrivacy training dataset size. FDF* shows DeepPrivacy trained for the full 40M (vs 4M) iterations, as reported in Fig.\ref{fig:general-comparison}. Dataset size is clearly important, but already for 38k PIPA images,  DeepPrivacy outperforms other GANs.}
    \label{fig:training-dataset-size}
\end{figure}

\textbf{Setup.} 
To test this hypothesis, we train DeepPrivacy on the PIPA dataset, just like Sun2018. Due to the simpler keypoints of DeepPrivacy, fewer images are disqualified from training and instead of 16,000 for Sun2018, this leaves 38,000 faces for training DeepPrivacy. To investigate the impact of training dataset size, we retrain DeepPrivacy for 4M iterations on the PIPA dataset with 38,000 faces, on a 100,000 faces subset of the FDF dataset~\cite{hukkelaas2019deepprivacy} and on the full 1.47M faces of FDF. DSFD is trained on the anonymized faces for 60,000 iterations. The results are shown in Fig.~\ref{fig:training-dataset-size}. As a frame of reference to other experiments, the scores for DSFD trained on original data and data anonymized by the fully trained DeepPrivacy of 40M iterations are also shown. Additionally, in this experiment we report the scores based on the easy, medium and hard validation sets of WIDER FACE, because intuitively a larger training set is especially necessary for the hard faces, which contain rare faces, such as strongly occluded ones and those with uncommon poses.

\begin{figure}[t]
    \centering
    \includegraphics[width=0.95\linewidth]{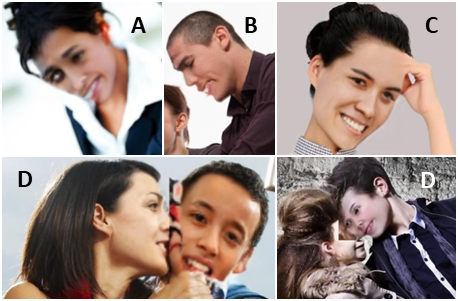}
    \caption{\textbf{Why DeepPrivacy performs best?} High-resolution failure cases of DeepPrivacy on specific types of ``easy" faces: (A) uncommon pose, (B) side view, (C) mild occlusion and (D) faces close together.}
    \label{fig:deepprivacy-failure-examples} 
\end{figure}

\textbf{Quantitative results.}
Fig.~\ref{fig:training-dataset-size} shows that the training dataset size of the anonymization GAN certainly matters, as expected. However, even with just 38,000 faces to train on, DeepPrivacy still reaches an average mAP of approximately 0.84, considerably outperforming the best scores of both Sun2018 and CIAGAN of previous experiments. Furthermore, moving from 100,000 to 1.47M training faces provides no further benefit, unless we also extend the number of training iterations from 4M to 40M. Surprisingly, when using a smaller dataset size, the drop in performance on ``easy" faces is the largest and on ``hard" faces the smallest. Investigating the generated faces visually reveals that specific cases of high-resolution ``easy" faces are the main reason. These cases include side-views of faces, faces close to other faces and mildly occluded faces, examples of which are shown in Fig.~\ref{fig:deepprivacy-failure-examples}. On low-resolution faces, these types of artefacts are far less common.
DeepPrivacy requires a large amount of training data to generate accurate high-resolution faces, while small faces (in ``hard"), look realistic when anonymized even with DeepPrivacy trained on few faces. Additionally, faces that are strongly occluded (also in ``hard") look poor regardless of the size of the dataset that DeepPrivacy is trained on.

\begin{figure}[t]
    \centering
    \includegraphics[width=0.95\linewidth]{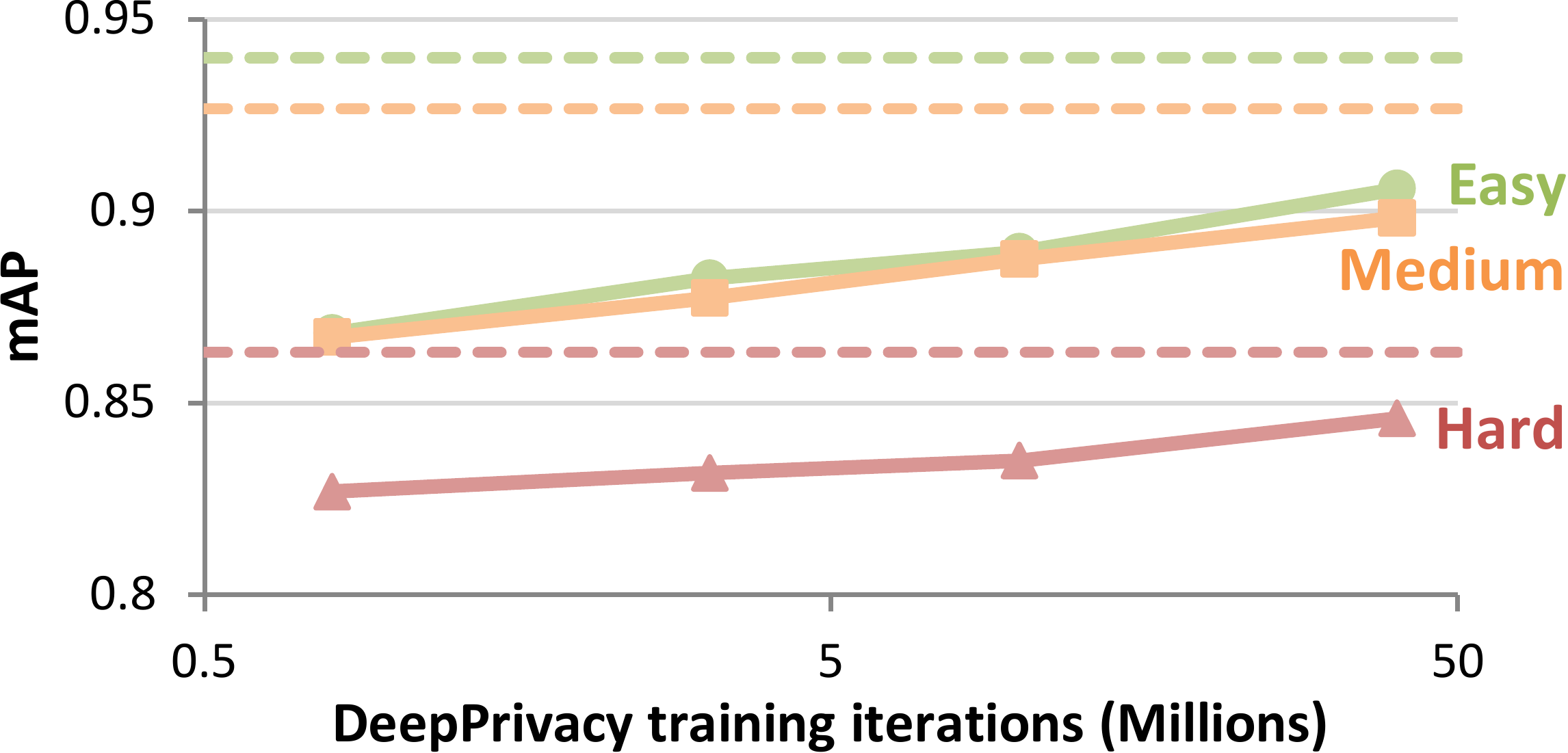}
    \caption{\textbf{Why DeepPrivacy performs best?} mAP scores of DSFD trained on data anonymized by DeepPrivacy as a function of the number of iterations. Training the anonymizer for more iterations keeps improving detector performance, with no clear sign of a plateau, even after millions of iterations.}
    \label{fig:anonymizer-training-iterations}
    \vspace{-3mm}
\end{figure}

In Fig.~\ref{fig:training-dataset-size}, there is a significant difference between the performance of DeepPrivacy trained for 4M iterations and DeepPrivacy trained for 40M iterations. To investigate this difference further, we train a detector on datasets anonymized with DeepPrivacy trained for 800,000 and 10M iterations and show the mAP in Fig.~\ref{fig:anonymizer-training-iterations}. Again, we plot the different WIDER FACE difficulties to visualize the effect on different types of faces. The largest difference in detector performance is again on the easy faces, verifying that faces that are easy to detect for a detector are actually hard to anonymize realistically. Another observation is that the performance improvement trend of training DeepPrivacy for more iterations is not clearly tapering off at 40M iterations, suggesting that training it for longer could still considerably improve performance. Sadly, the next point on this logarithmic scale is 400 million iterations, which would take approximately 400 days to train on our machine, which is unfeasible to evaluate.

\subsection{What GAN metric correlates with face detector mAP?}
\textbf{Setup.} Training a face detector is time-consuming, hence it is useful to know if common metrics used for evaluating GANs correlate well with the performance of face detectors trained on anonymized data. A metric with high correlation could be used as an inexpensive alternative to full detector training during anonymization experiments. To verify this, we take all anonymized datasets used for the experiments in this paper and compute five metrics that can be employed to evaluate image-inpainting GANs. These metrics are L1 and L2 distance, Learned Perceptual Image Patch Similarity (LPIPS)~\cite{zhang2018perceptual}, Fr\'echet Inception Distance (FID)~\cite{heusel2017gans}, and Structural Similarity Index Measure (SSIM)~\cite{Wang2004}, measured between original and anonymized faces. The L1 and L2 distances are easy to compute, even at train time, and often actively minimized via a reconstruction loss in the network. LPIPS measures visual similarity by comparing images in feature space of an independent ImageNet-trained classifier. FID is similar to LPIPS, but measures the distance between distributions of images instead of individual images, making it most suitable to test for both realism and variety of generated faces, but is more expensive to compute. Finally, SSIM compares luminance, contrast and structure to estimate human perceptual difference.

\textbf{Quantitative results.} The correlation of these metrics with the achieved mAP of training a detector on the corresponding anonymized images is shown in Table~\ref{tab:correlations}. Overall FID appears most suitable with an absolute correlation of 0.78 over all experiments (labeled ``All"), but 0.78 is not sufficient to view the metric as a replacement to actual detector training on anonymized data. However, it appears that all tested metrics simply behave very differently for conventional methods rather than for GANs, because splitting the correlation computation between GAN and conventional yields vastly higher correlation values, as shown in the second and third column of Table ~\ref{tab:correlations}. To conclude, although these metrics are not suited for comparing arbitrary anonymization methods, they are useful for comparing different GAN-based methods. Considering that recent GANs vastly outperform conventional methods, the metrics remain suitable as an alternative to training a face detector on anonymized data.

\setlength{\tabcolsep}{5pt} 
\renewcommand{\arraystretch}{1.1} 
\begin{table}[t]
\caption{\textbf{What GAN metric correlates with face detector mAP?}
Correlations of common GAN evaluation metrics with the achieved mAP score of DSFD of previous experiments. All metrics are suited for comparing GANs, but none for comparing arbitrary anonymization methods.}
\label{tab:correlations}
\centering
\begin{tabular}{lccc}
\hline
\textbf{Correlation with:} & \textbf{All} & \textbf{GAN} & \textbf{Conventional} \\ \hline
FID                        & -0.777       & -0.978       & -1.000                \\
L1                         & -0.458       & -0.996       & -0.999                \\
L2                         & -0.454       & -0.994       & -0.999                \\
LPIPS                      & -0.665       & -0.992       & -0.971                \\
SSIM                       & 0.721        & 0.983        & 0.759                 \\ 
\hline
\end{tabular}
\end{table}


\section{Discussion}

The minimum face resolution required for anonymization to be applicable is around 14 pixels wide for DeepPrivacy and around 50 pixels wide for CIAGAN and Sun2018. However, there are no regulations or guidelines for the minimum size of a face to be considered privacy sensitive. A minimum size required to identify a face should be determined in future research, as it has not been explored in this paper. 

When using two different datasets for training the anonymizer and the face detector, the domain gap of the datasets should be certainly considered. This gap is caused by (distribution) differences between the datasets, such as face size, pose distribution and context type. Even in the best case, it cannot be expected that the generated faces that were learned from other datasets, will perfectly match the target domain. Fine-tuning the anonymizer on the target domain, which is the simplest method of dealing with domain shift, could further improve the anonymizer.

Future work could look deeper into selectively anonymizing datasets. Anonymization GANs perform especially poorly on odd poses and heavily occluded faces, but the latter category is already difficult to recognize. Intelligently selecting a set of long-tail distribution faces not to anonymize, could increase detector performance significantly, while still anonymizing the vast majority of faces in the dataset. 

\newpage
\section{Conclusion}
This work attempts to answer the research question: ``Can we anonymize faces while maintaining effective training of face detectors?" Anonymizing the training data of face detectors using conventional methods such as blurring, blocking or pixelation, results in a very poor detection performance. Anonymization using modern face generation GANs is possible and gives good visual results. We are the first to investigate the impact of this anonymization on the performance of training a DSFD face detector. DeepPrivacy~\cite{hukkelaas2019deepprivacy} outperforms other GAN-based methods, resulting in only 2.7\% mAP degradation for the face detector, compared to training on original non-anonymized data.

We have investigated the effectiveness and underlying motivation of anonymizers in four experiments. A good anonymization method introduces few artefacts and as a result, the longer a detector is trained on the anonymized data, the higher its performance. Methods that introduce significant artefacts, such as conventional methods and Sun2018, cause face detectors to overtrain on the artefacts, instead of learning general face appearance. Sadly, even for DeepPrivacy, face detectors trained on anonymized data suffer an increasing penalty to mAP when the detector is trained longer (1.6\% mAP at 7,500 iterations, to 2.7\% mAP at 60,000  iterations). This suggests that the penalty for anonymizing data will increase as better detectors are developed, unless face anonymization networks improve as well.


Finally, we have explored standard metrics that are used for evaluating GANs and whether these metrics correlate well with the mAP score of a detector trained on anonymized data. All five tested metrics, L1 distance, L2 distance, LPIPS, FID and SSIM, correlate well with training performance when mutually comparing different GANs, but not when comparing other methods with GANs. Hence, when developing GANs for anonymization, these standard metrics can be used to estimate face detector performance.

Although the separate fields of face detection and face anonymization have been researched extensively, their combination opens up more research questions. For example, high-resolution faces are easy to detect but hard to anonymize without artefacts, leading to interesting mutual relationships for future research.

{\small
\bibliographystyle{ieee}
\bibliography{references.bib}

\begin{thebibliography}{10}\itemsep=-1pt

\bibitem{Cho2020}
D.~Cho, J.~H. Lee, and I.~H. Suh.
\newblock {CLEANIR: Controllable attribute-preserving natural identity
  remover}.
\newblock {\em Applied Sciences}, 10(3), 2020.

\bibitem{deng2019retinaface}
J.~Deng, J.~Guo, Y.~Zhou, J.~Yu, I.~Kotsia, and S.~Zafeiriou.
\newblock {Retinaface: Single-stage dense face localisation in the wild}.
\newblock {\em arXiv:1905.00641}, 2019.

\bibitem{earp2019face}
S.~W.~F. Earp, P.~Noinongyao, J.~A. Cairns, and A.~Ganguly.
\newblock {Face Detection with Feature Pyramids and Landmarks}.
\newblock {\em arXiv:1912.00596}, 2019.

\bibitem{goodfellow2014generative}
I.~Goodfellow, J.~Pouget-Abadie, M.~Mirza, B.~Xu, D.~Warde-Farley, S.~Ozair,
  A.~Courville, and Y.~Bengio.
\newblock {Generative adversarial nets}.
\newblock In {\em NIPS}, 2014.

\bibitem{Guo2016}
Y.~Guo, L.~Zhang, Y.~Hu, X.~He, and J.~Gao.
\newblock {MS-celeb-1M: A dataset and benchmark for large-scale face
  recognition}.
\newblock In {\em Electronic Imaging, Imaging and Multimedia Analytics in a Web
  and Mobile World}, 2016.

\bibitem{He}
K.~He, G.~Gkioxari, P.~Doll, and R.~Girshick.
\newblock {Mask R-CNN}.
\newblock In {\em ICCV}, 2017.

\bibitem{heusel2017gans}
M.~Heusel, H.~Ramsauer, T.~Unterthiner, B.~Nessler, and S.~Hochreiter.
\newblock {Gans trained by a two time-scale update rule converge to a local
  nash equilibrium}.
\newblock In {\em NIPS}, 2017.

\bibitem{hu2017finding}
P.~Hu and D.~Ramanan.
\newblock {Finding tiny faces}.
\newblock In {\em CVPR}, 2017.

\bibitem{hukkelaas2019deepprivacy}
H.~Hukkel{\aa}s, R.~Mester, and F.~Lindseth.
\newblock {Deepprivacy: A generative adversarial network for face
  anonymization}.
\newblock In {\em International Symposium on Visual Computing}, pages 565--578.
  Springer, 2019.

\bibitem{JiaDeng2009}
{Jia Deng}, {Wei Dong}, R.~Socher, {Li-Jia Li}, {Kai Li}, and {Li Fei-Fei}.
\newblock {ImageNet: A large-scale hierarchical image database}.
\newblock {\em CVPR}, 2009.

\bibitem{karras2019style}
T.~Karras, S.~Laine, and T.~Aila.
\newblock {A style-based generator architecture for generative adversarial
  networks}.
\newblock In {\em CVPR}, 2019.

\bibitem{kundur1996blind}
D.~Kundur and D.~Hatzinakos.
\newblock {Blind image deconvolution}.
\newblock {\em IEEE signal processing magazine}, 13(3):43--64, 1996.

\bibitem{li2019dsfd}
J.~Li, Y.~Wang, C.~Wang, Y.~Tai, J.~Qian, J.~Yang, C.~Wang, J.~Li, and
  F.~Huang.
\newblock {DSFD: Dual shot face detector}.
\newblock In {\em CVPR}, 2019.

\bibitem{liu2016ssd}
W.~Liu, D.~Anguelov, D.~Erhan, C.~Szegedy, S.~Reed, C.-Y. Fu, and A.~C. Berg.
\newblock {SSD: Single shot multibox detector}.
\newblock In {\em ECCV}, 2016.

\bibitem{liu2015faceattributes}
Z.~Liu, P.~Luo, X.~Wang, and X.~Tang.
\newblock {Deep Learning Face Attributes in the Wild}.
\newblock In {\em ICCV}, 2015.

\bibitem{ma2017pose}
L.~Ma, X.~Jia, Q.~Sun, B.~Schiele, T.~Tuytelaars, and L.~Van~Gool.
\newblock {Pose guided person image generation}.
\newblock In {\em NIPS}, 2017.

\bibitem{maximov2020ciagan}
M.~Maximov, I.~Elezi, and L.~Leal-Taix{\'{e}}.
\newblock {CIAGAN: Conditional Identity Anonymization Generative Adversarial
  Networks}.
\newblock In {\em CVPR}, 2020.

\bibitem{Murgia2019}
M.~Murgia.
\newblock {Microsoft quietly deletes largest public face recognition data set}.
\newblock In {\em Financial Times}, 2019.

\bibitem{newton2005preserving}
E.~M. Newton, L.~Sweeney, and B.~Malin.
\newblock {Preserving privacy by de-identifying face images}.
\newblock {\em IEEE transactions on Knowledge and Data Engineering},
  17(2):232--243, 2005.

\bibitem{padilla2015visual}
J.~R. Padilla-L{\'{o}}pez, A.~A. Chaaraoui, and F.~Fl{\'{o}}rez-Revuelta.
\newblock {Visual privacy protection methods: A survey}.
\newblock {\em Expert Systems with Applications}, 42(9):4177--4195, 2015.

\bibitem{ren2015faster}
S.~Ren, K.~He, R.~Girshick, and J.~Sun.
\newblock {Faster R-CNN: Towards real-time object detection with region
  proposal networks}.
\newblock In {\em NIPS}, 2015.

\bibitem{ribaric2016identification}
S.~Ribaric, A.~Ariyaeeinia, and N.~Pavesic.
\newblock {De-identification for privacy protection in multimedia content: A
  survey}.
\newblock {\em Signal Processing: Image Communication}, 47:131--151, 2016.

\bibitem{Ristani2016}
E.~Ristani, F.~Solera, R.~Zou, R.~Cucchiara, and C.~Tomasi.
\newblock {Performance measures and a data set for multi-target, multi-camera
  tracking}.
\newblock In {\em ECCV workshop on Benchmarking Multi-Target Tracking}, 2016.

\bibitem{Ronneberger2015}
O.~Ronneberger, P.~Fischer, and T.~Brox.
\newblock {U-Net: Convolutional networks for biomedical image segmentation}.
\newblock In {\em Medical Image Computing and Computer-Assisted Intervention},
  volume 9351, 2015.

\bibitem{Satisky2019}
J.~Satisky.
\newblock {A Duke study recorded thousands of students’ faces. Now they’re
  being used all over the world}.
\newblock In {\em The Chronicle}, 2019.

\bibitem{sun2018natural}
Q.~Sun, L.~Ma, S.~Joon~Oh, L.~Van~Gool, B.~Schiele, and M.~Fritz.
\newblock {Natural and effective obfuscation by head inpainting}.
\newblock In {\em CVPR}, 2018.

\bibitem{sun2018hybrid}
Q.~Sun, A.~Tewari, W.~Xu, M.~Fritz, C.~Theobalt, and B.~Schiele.
\newblock {A hybrid model for identity obfuscation by face replacement}.
\newblock In {\em ECCV}, pages 553--569, 2018.

\bibitem{Thomee_2016}
B.~Thomee, D.~A. Shamma, G.~Friedland, B.~Elizalde, K.~Ni, D.~Poland, D.~Borth,
  and L.-J. Li.
\newblock {YFCC100M}.
\newblock {\em Communications of the ACM}, 59(2):64–73, 1 2016.

\bibitem{Torralba2008}
A.~Torralba, R.~Fergus, and W.~T. Freeman.
\newblock {80 million tiny images: a large data set for nonparametric object
  and scene recognition}.
\newblock {\em IEEE Transactions on Pattern Analysis and Machine Intelligence},
  30(11):1958--1970, 2008.

\bibitem{voigt2017eu}
P.~Voigt and A.~dem Bussche.
\newblock {The EU General Data Protection Regulation (GDPR)}.
\newblock {\em A Practical Guide, 1st Ed., Cham: Springer International
  Publishing}, 2017.

\bibitem{Wang2004}
Z.~Wang, A.~C. Bovik, H.~R. Sheikh, and E.~P. Simoncelli.
\newblock {Image quality assessment: From error visibility to structural
  similarity}.
\newblock {\em IEEE Transactions on Image Processing}, 13(4):600--612, 2004.

\bibitem{Yang2021}
K.~Yang, J.~Yau, L.~Fei-Fei, J.~Deng, and O.~Russakovsky.
\newblock {A Study of Face Obfuscation in ImageNet}.
\newblock {\em ArXiv: 2103.06191v2}, 2021.

\bibitem{yang2016wider}
S.~Yang, P.~Luo, C.~C. Loy, and X.~Tang.
\newblock {WIDER FACE: A Face Detection Benchmark}.
\newblock In {\em CVPR}, 2016.

\bibitem{zhang2020asfd}
B.~Zhang, J.~Li, Y.~Wang, Y.~Tai, C.~Wang, J.~Li, F.~Huang, Y.~Xia, W.~Pei, and
  R.~Ji.
\newblock {ASFD: Automatic and Scalable Face Detector}.
\newblock {\em arXiv preprint arXiv:2003.11228}, 2020.

\bibitem{zhang2019accurate}
F.~Zhang, X.~Fan, G.~Ai, J.~Song, Y.~Qin, and J.~Wu.
\newblock {Accurate face detection for high performance}.
\newblock {\em arXiv:1905.01585}, 2019.

\bibitem{piper}
N.~Zhang, M.~Paluri, Y.~Taigman, R.~Fergus, and L.~Bourdev.
\newblock {Beyond Frontal Faces: Improving Person Recognition Using Multiple
  Cues}.
\newblock In {\em CVPR}, 2015.

\bibitem{zhang2018perceptual}
R.~Zhang, P.~Isola, A.~A. Efros, E.~Shechtman, and O.~Wang.
\newblock {The Unreasonable Effectiveness of Deep Features as a Perceptual
  Metric}.
\newblock In {\em CVPR}, 2018.

\end{thebibliography}
}

\end{document}